\documentclass{article}

\usepackage{arxiv}
\usepackage[utf8]{inputenc} 
\usepackage[T1]{fontenc}    
\usepackage{hyperref}       
\usepackage{url}            
\usepackage{booktabs}       
\usepackage{amsfonts}       
\usepackage{nicefrac}       
\usepackage{microtype}      

\usepackage{graphicx}
\usepackage{doi}
\usepackage{wrapfig}

\usepackage{diagbox}
\usepackage{makecell}
\usepackage{multirow}
\usepackage{amsmath,amssymb}
\usepackage{amsthm}
\usepackage{mathrsfs}
\usepackage{float}
\usepackage[justification=centering]{caption}

\usepackage[numbers, sort&compress]{natbib}
\setcitestyle{numbers}

\title{Wear Classification of Abrasive Flap Wheels using a Hierarchical Deep Learning Approach}

\usepackage{authblk}

\newbox{\orcid}\sbox{\orcid}{\includegraphics[scale=0.06]{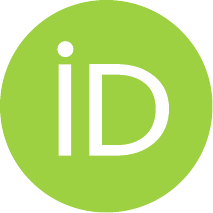}} 
\author[1]{%
   \href{https://orcid.org/0000-0003-4732-8364}{\usebox{\orcid}\hspace{1mm}Falko Kähler\thanks{\texttt{f.kaehler@tuhh.de}}}%
}
\author[1]{%
   \href{https://orcid.org/0009-0000-2335-4475}{\usebox{\orcid}\hspace{1mm}Maxim Wille\thanks{\texttt{maxim.wille@tuhh.de}}}%
}
\author[1]{%
   \href{https://orcid.org/0000-0001-6488-5141}{\usebox{\orcid}\hspace{1mm}Ole Schmedemann\thanks{\texttt{ole.schmedemann@tuhh.de}}}%
}
\author[1]{%
   \href{https://orcid.org/0000-0002-9616-3976}{\usebox{\orcid}\hspace{1mm}Thorsten Schüppstuhl\thanks{\texttt{schueppstuhl@tuhh.de}}}%
}

\affil[1]{Hamburg University of Technology, Institute of Aircraft Production Technology, 21073 Hamburg, Germany.}

\renewcommand{\shorttitle}{Wear Classification of Abrasive Flap Wheels}

\hypersetup{
pdftitle={Wear Classification of Abrasive Flap Wheels using a Hierarchical Deep Learning Approach},
pdfsubject={cs.CV, eess.IV},
pdfauthor={First Author, Second Author, Third Author},
pdfkeywords={abrasive flap wheel, wear, classification},
}

\begin{document}
\maketitle

\begin{abstract}
Abrasive flap wheels are common for finishing complex free-form surfaces due to their flexibility. However, this flexibility results in complex wear patterns such as concave/convex flap profiles or flap tears, which influence the grinding result. This paper proposes a novel, vision-based hierarchical classification framework to automate the wear condition monitoring of flap wheels. Unlike monolithic classification approaches, we decompose the problem into three logical levels: (1) state detection (new vs. worn), (2) wear type identification (rectangular, concave, convex) and flap tear detection, and (3) severity assessment (partial vs. complete deformation). A custom-built dataset of real flap wheel images was generated and a transfer learning approach with EfficientNetV2 architecture was used. The results demonstrate high robustness with classification accuracies ranging from 93.8\% (flap tears) to 99.3\% (concave severity). Furthermore, Gradient-weighted Class Activation Mapping (Grad-CAM) is utilized to validate that the models learn physically relevant features and examine false classifications. The proposed hierarchical method provides a basis for adaptive process control and wear consideration in automated flap wheel grinding.
\end{abstract}


\section{Introduction}
Abrasive flap wheels are widely used in manufacturing for finishing tasks of complex geometries due to their flexibility and high achievable surface finish, e.g. in aerospace industry \citep{Zhang.2017, Kaehler.2023} or mold making \citep{Wilbert.2015}. However, manual grinding requires skilled and experienced staff due to the complex behavior of flap wheels, which in turn limits reproducibility of grinding results and overall productivity. To meet demand for increased productivity and quality in future, automation of the manual flap wheel grinding process is essential \citep{Kaehler.2023, Chen.2019, Zhang.2022}.

Yet, the flap wheel wear poses challenges, as it affects the tool behavior and grinding result, leading to inconsistent grinding results, machining errors or even scrapping of valuable workpieces if not incorporated. While the experienced worker can visually assess the flap wheel condition and compensate or avoid wear effects, an automation system must visually inspect the tool and autonomously determine its condition. For flap wheels, the challenge lies in the high variance of geometric wear patterns resulting from different contact situations and process parameters. Consequently, classical rule-based image processing methods, which are ultimately necessary to measure wear precisely, may be prone to errors when applied standalone across this wide range of wear patterns.

Addressing this issue, this paper presents an AI-driven approach designed to enhance the robustness of geometric flap wheel wear analysis. By introducing a hierarchical classification framework that independently assesses wear, potential ambiguities can be resolved and measurement results of dedicated image processing approaches validated, providing a solid decision-making basis for adaptive process control.

This paper is structured as follows: In section \ref{sec:StateoftheArt}, flap wheels and research on their wear is elaborated, as well as methods of tool wear condition monitoring. Section \ref{sec:WearAnalysisAndApproach} describes the geometric wear patterns of abrasive flap wheels, then a classification approach is selected and the hierarchical classification approach is presented. Section \ref{sec:Implementation} covers data generation for the AI models as well as implementation details, while section \ref{sec:Results} presents the classification results for each network as well as the complete hierachical classification. Finally, the conclusions of this research are drawn in section \ref{sec:Conclusion}.

\section{State of the Art} \label{sec:StateoftheArt}
\subsection{Wear of abrasive flap wheels}
\begin{figure}[htb]
\centering
\includegraphics{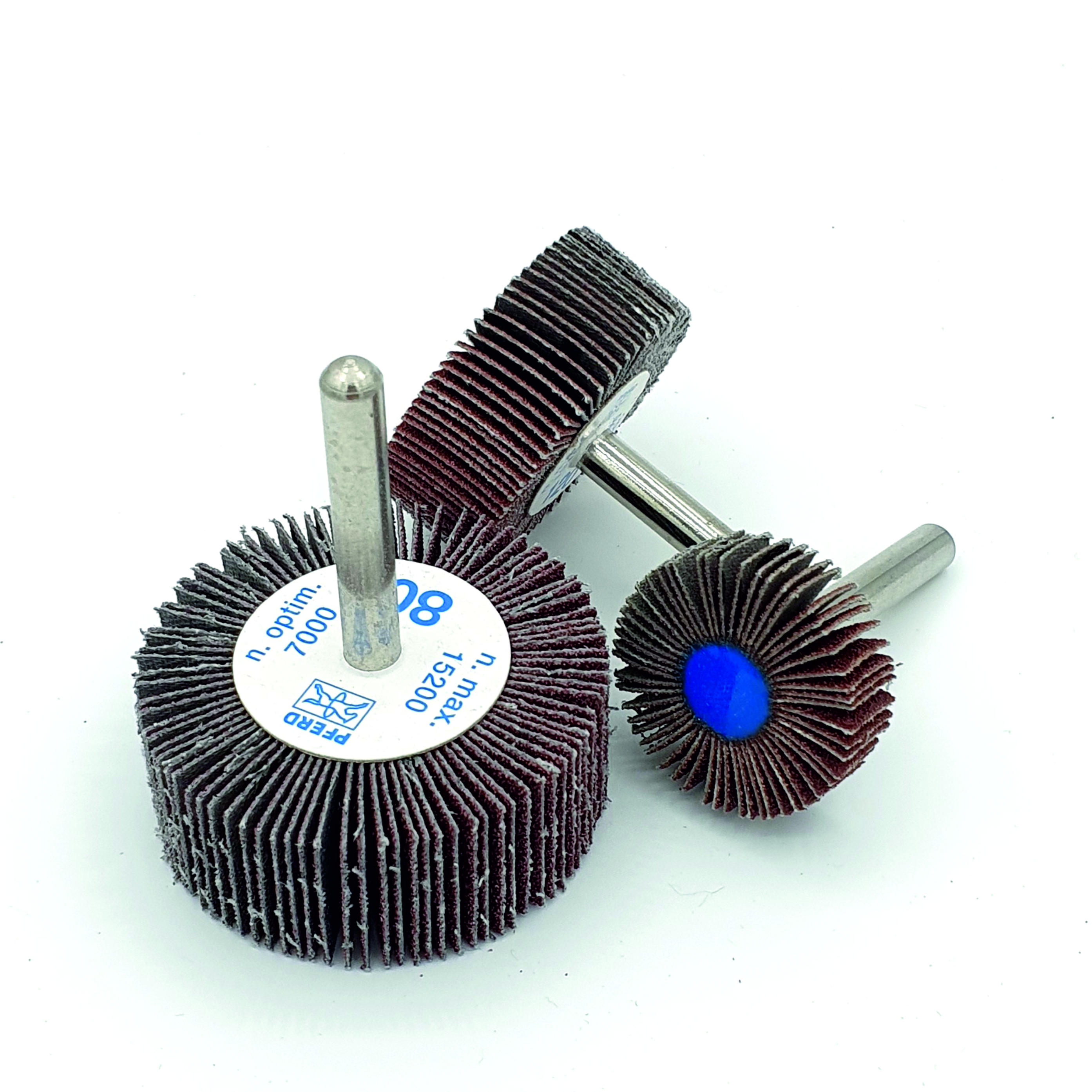}
\caption{Abrasive flap wheels \citep{Kaehler.2025} \label{fig:FlapWheels}}
\end{figure}

Abrasive flap wheels, as seen in Figure \ref{fig:FlapWheels}, consist of radially mounted abrasive flaps on a plastic core, fixed on a metal shaft. The flaps consist of a substrate (usually textile fabric) to which the abrasive grains are adhered using a binder. This structure corresponds to coated abrasives and gives the flap wheel its flexibility and shape adaptability \citep{Zhu.2020, Zhang.2017}.

While material removal modeling \citep{Lin.2019, Zheng.2021} or path planning \citep{Chen.2024, Huai.2019, Zhang.2024} are already covered in various research, wear is usually not taken into account and circumvented by changing the tool regularly and prematurely, leading to restrictions in use or inefficient tool utilisation. Research focussed explicitly on flap wheel wear, however, is limited. \citet{Toenshoff.1985} investigated unmounted flap wheels during outer circumferential grinding and examined the flap length degradation. The authors described a initial sharpness and tapering of the flap edges at the very first use. After loss of the initial sharpness, wear of abrasive grit, binder and substrate is uniform.

The progressive flap length degradation, however, leads to changing technological parameters, i.e. the effective cutting velocity and workpiece overlap decrease with decreasing diameter. In order to compensate the wear, an increased rotation speed and adjusted contact force method is proposed. However, in terms of uniform roughness and material removal, a constant cutting speed and contact force were found to be beneficial. The authors also stated the option to dress flap wheels, however, this would decrease efficient tool use.

This initial sharpness described by \citet{Toenshoff.1985} is confirmed by \citet{Yang.2024}, where the authors focused on the grit wear and identified three wear stages, with different proportions of wear mechanisms (blunt, stepped fracture, cleavage fracture, pullout and adhesive wear). Interestingly, the authors observed an increased wear rate with increasing spindle speed, which in combination with the approach of \citet{Toenshoff.1985} might progressively increase wear rate. However, the overall wear amount was relatively low, so that it is unclear whether the findings are transferable at late stages of flap wheel wear.

\citet{Kaehler.2025} investigated geometric flap wheel wear. Focussing on concave flap profiles and their effect on material removal uniformity, it became evident the geometric wear has significant influence on the grinding result. A compensation method was demonstrated by adjusting the flap wheel orientation during robotic grinding. However, the necessary path adjustment depended on the severity of flap concavity.

\subsection{Wear condition monitoring}
Wear condition monitoring in research and industrial practice relies either on indirect process signals or direct, usually vision-based methods. For direct monitoring, a review of the current literature reveals that AI-driven approaches — particularly Convolutional Neural Networks (CNNs) — are increasingly preferred over rule-based image processing, especially when image content, lighting conditions, or tool geometries exhibit high variability \citep{Pimenov.2023, Cheng.2019}.

AI wear condition monitoring usually involves categorizing tools into defined states, such as "new" versus "worn," or identifying specific wear mechanisms. For instance, \citet{Wu.2019} employed a pre-trained CNN to classify milling tools into four categories (including adhesion, breakage, and flank wear), achieving an average accuracy of 96.2\% . \citet{Wolf.2025} applied several image-based classification algorithms to classify the coating of cutting tools into "new" or "worn". In their research, decision trees showed best performance achieving a F1-score of 0.95, indicating effective wear detection. \citet{Oo.2020} determined the wear state of abrasive belts by using the abrasive grain area and random forest classification to distinguish three wear periods.

Alternatively, indirect methods for assessing wear conditions are often employed. These methods utilize process data such as forces, vibrations, temperatures, sound or machining results, which highly correlate with wear states \citep{Pandiyan.2020, Pimenov.2025}. For instance, \citet{Kumar.2021} and \citet{Sujay.2025} analyzed the texture of the machined metallic surface, both reporting wear classification accuracies of up to 99.9\%. \citet{Surindra.2025} applied several machine learning methods such as Support Vector Machines or Decision Trees on accelerometer and force sensor signals to determine the wear state during robotic belt grinding.

While indirect monitoring methods, such as the surface texture or process signal analysis, demonstrate high robustness for specific applications, they face inherent limitations regarding flexibility. When applied to the finishing of complex free-form surfaces, the reliability depends heavily on accessibility of the machined surface as well as consistent acquisition situations, e.g. sensor positioning, lighting angles or focal distances relative to the workpiece surface. Adapting to varying geometries would require elaborate, component-specific sensor positioning or path planning. Consequently, the engineering effort required to adapt the sensor system makes indirect monitoring economically viable primarily for high-volume production of low-complexity parts. Furthermore, indirect surface inspection is often reactive, identifying defects only after the workpiece has potentially been machined incorrectly. Indirect methods therefore should be coupled with a tool life prediction.

In contrast, direct tool inspection provides insight into the specific wear position or extent prior to the next process step. This effectively decouples the monitoring task from the workpiece geometry. By using a dedicated inspection pose within a static sensor setup, the system ensures robust data acquisition regardless of component complexity. This independence indicates the direct approach is a superior enabler for the automated finishing of complex parts.

\subsection{Conclusion}
A review of the scientific literature reveals a gap in the domain of automated wear monitoring for abrasive flap wheels. While direct vision-based wear monitoring is a well-established field for rigid tools such as milling cutters \citep{Wu.2019} or turning inserts \citep{Wolf.2025}, flap wheels and their wear characteristics have received comparatively little attention. This disparity is primarily due to the tool's inherent flexibility, which results in fundamentally different, non-linear wear behavior. Unlike rigid tools that typically exhibit predictable wear, flap wheels are subject to complex geometric deformations such as concave or convex profile changes, and stochastic structural defects like flap tears.

Consequently, conventional rule-based image processing methods face inherent limitations in capturing the high variance of these wear patterns. Moreover, current research has demonstrated successful application of data-driven methods, particularly Convolutional Neural Networks (CNNs), in capturing complex surface textures and irregular defect patterns in other manufacturing domains. These capabilities suggest that deep learning models are well-suited to address the high variance of flap wheel wear. However, a specific framework that addresses the wear characteristics of flap wheels is currently lacking.

\section{Wear analysis and classification approach}\label{sec:WearAnalysisAndApproach}
\subsection{Wear analysis}\label{sec:WearAnalysis}
To identify wear patterns, several flap wheels with different wear conditions and use times were analyzed in the laboratory environment and at industrial users. It was observed that used flap wheels differ from new ones not only in obvious signs of geometric wear, but also in terms of lacking textile fringes and a grayish discoloration of the flap edge, which is the underlying fabric base showing through.

With increasing usage time, abrasion of the abrasive and the backing occurs starting from the radial edge of the flap, so that the flap length and thus the tool diameter gradually decrease. If the flaps shorten evenly across their width, the rectangular flap shape is retained, whereas uneven flap shortening due to uneven flap load leads to a changed flap profile. The workpiece contour is roughly reproduced in the flap as a result of the uneven load. This profiling can take place either over parts or over the entire flap width and is referred to below as partial or complete profiling.

The flap profile is thus the result of the superimposition of the engagement situations and the resulting pressure distributions over the operating time, leading to a variety of possible flap profiles. Theoretically, partial concavity and convexity are possible in parallel, especially with wide flap wheels. In practice, however, this rarely occurs and is therefore not taken into account here.

\begin{figure}[htb]
\centering
\includegraphics[width=10.0 cm]{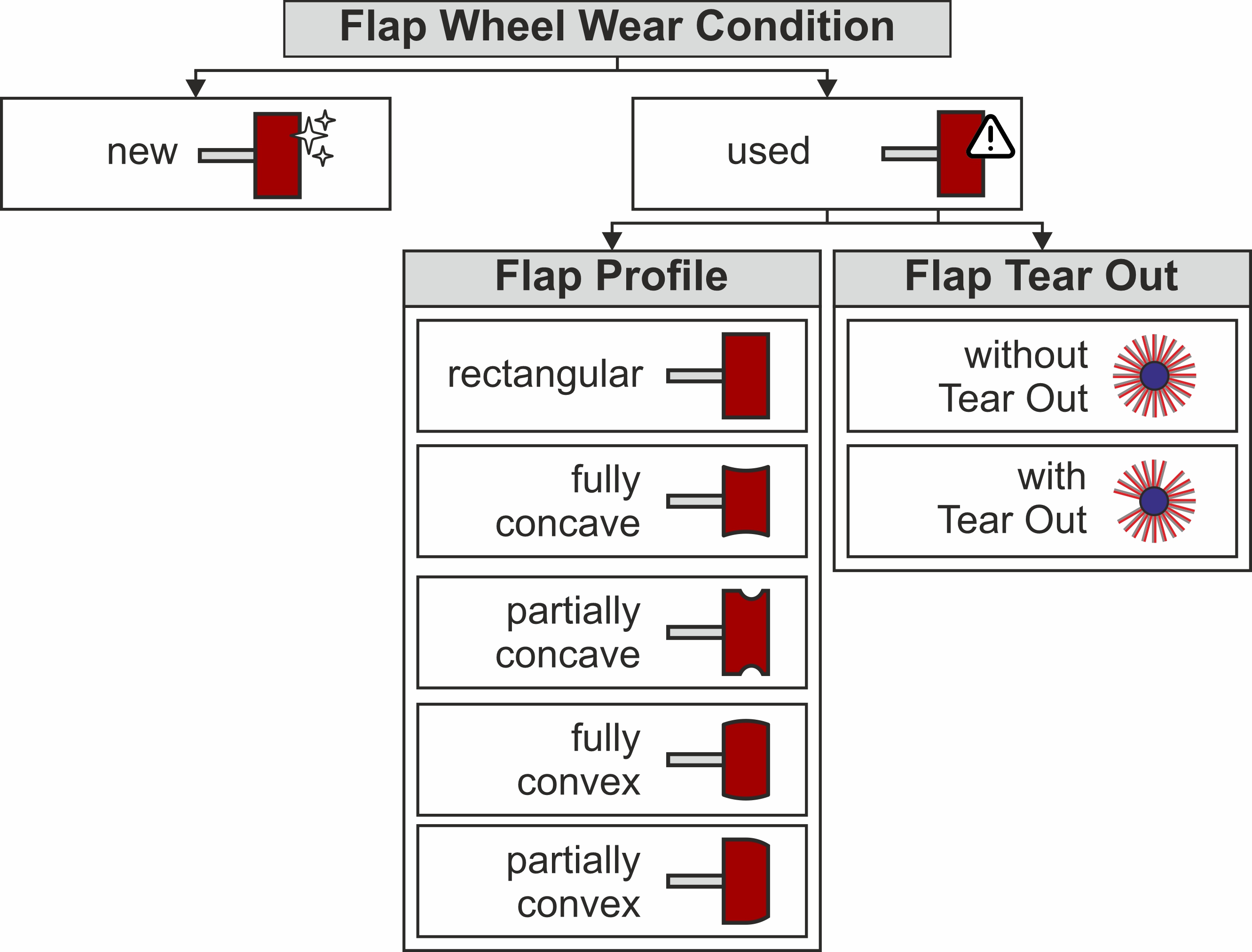}
\caption{Flap wheel wear categorization\label{fig:WearCategories}}
\end{figure}

Parallel to flap length degradation and/or flap profiling, total tear out of flaps can occur, leaving only flap stubs at the core. Tear out increases the space between flaps and thus local tool flexibility, since their mutual support decreases. However, when it comes to detecting flap tears, the gaps are not a sufficient indicator, as the gaps always vary slightly. Image processing approaches analyzing the gaps therefore would fail and have to detect the flap stubs, which in turn is not feasible due to variations in texture and color of the flaps. Figure \ref{fig:WearCategories} depicts the identified wear categories.

\subsection{Classification Approach Selection}
Based on the geometric flap wheel wear patterns identified in section \ref{sec:WearAnalysis}, it is clear that these can be diverse, requiring a correspondingly robust classification. There are various approaches to implementing image-based categorization of tools. In comparison to direct multi-class classification, in which all states/classes are distinguished in a single model, hierarchical classification represents a particularly suitable alternative as it divides the overall task into several consecutive sub-decisions \citep{Narushynska.2025, Cerri.2014, Silla.2011}.

These sub-decisions can be assigned to individual wear characteristics. This decomposition reduces the necessary selectivity of each network, since each subnetwork can be individually trained for a specific task \citep{Yan.2014}. A certain modularity of the networks is enabled, e.g. for further refinement/extension of individual networks, and avoids combinatorial data fragmentation. While a direct multi-class classifier requires a statistically significant amount of training images for each possible wear combination, the hierarchical approach considers these features independently.

This is particularly relevant for the detection of rare events: In a flat structure, the few available images of defects would be divided into different geometric subclasses, which would result in the respective class basis being too small for robust training. Hierarchical decoupling, on the other hand, allows all images showing a tear—regardless of their other characteristics—to be used to train the flap tear network. This increases the amount of data per learned feature and enables robust detection even with a smaller overall data set.

The hierarchical structure also allows logical dependencies between classes to be taken into account, so that any errors can be detected. Compared to a multi-class classifier, this improves robustness against classification errors, as logical inconsistencies between successive decisions can serve as an indication of misclassifications \citep{Narushynska.2025, Silla.2011}.

Another advantage is that different viewing perspectives can be used specifically for different partial decisions and networks. For example, flap tears are best detected when viewing the front of the flap disc axially, while the flap shape can only be assessed from a radial perspective. On the other hand, a disadvantage of hierarchical classification compared to a single, large classifier can be longer processing times due to the sequential sequence. However, given today's computing capacities, this is not considered a significant disadvantage here. Because of the advantages of hierarchical classification over direct multi-class classification, this approach is being pursued.

\subsection{Hierarchical Classification Design}
\begin{figure}[H]
\centering
\includegraphics[width=10.0 cm]{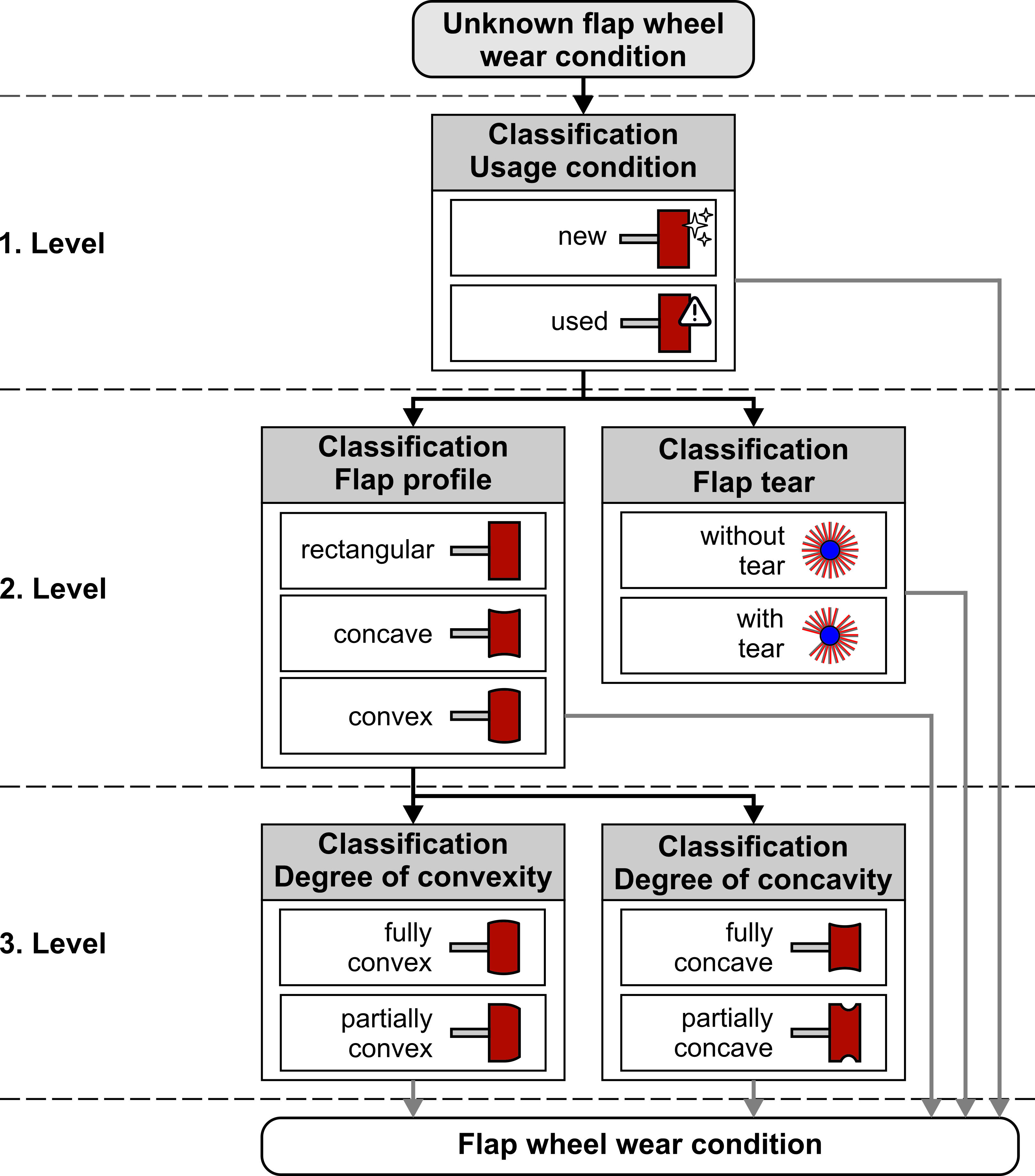}
\caption{Hierarchical wear classification design\label{fig:Hierarchical_wear_classification}}
\end{figure}

The subtasks for wear patterns were linked to a three-level hierarchical classification with increasing levels of detail, as shown in Figure \ref{fig:Hierarchical_wear_classification}. The first level involves clarifying the usage condition (new or used). This is followed in level 2 by determining the flap profile (rectangular, concave or convex) and identifying flap tears (existent / non-existent), while in level 3, after prior classification as concave or convex, further detailing of the profile is carried out in fully or partially.

Possible paths in the hierarchy tree can be divided into consistent or contradictory results using logical checks. All consistent results are recorded in Table \ref{tab:Ergebniskategorien}. The logical consistency check is performed in particular between level 1 and 2. Possible conflicts are summarized in Table \ref{tab:Konfliktkategorien}. For example, a condition classification of “new” contradicts the classifications “with flap tear” or flap profile "concave" or “convex”\footnote{This study assumes a new flap wheel has a rectangular flap shape. Therefore, this contradiction is context-dependent as there are spherical shaped flap wheels available.}.

\begin{table}[htb]
\centering
\caption{Consistent results of the hierarchical wear classification}\label{tab:Ergebniskategorien}
\begin{tabular}{|c|c|cc|cc|}
\hline 
\textbf{Hierarchy level} & \textbf{1} & \multicolumn{2}{c|}{\textbf{2}}&\multicolumn{2}{c|}{\textbf{3}} \\
\hline 
\textbf{No.} & \textbf{Use condition} & \textbf{Flap profile} & \textbf{Flap tear} & \textbf{concave} & \textbf{convex} \\
\hline
\hline
1 & new & rectangular & no tear & n.a. & n.a. \\
\hline 
2 & used & rectangular & no tear & n.a. & n.a. \\
\hline 
3 & used & rectangular & with tear & n.a. & n.a. \\
\hline 
4 & used & concave & no tear & partially & n.a. \\
\hline 
5 & used & concave & with tear & partially & n.a. \\
\hline 
6 & used & concave & no tear & fully & n.a. \\
\hline 
7 & used & concave & with tear & fully & n.a. \\
\hline 
8 & used & convex & no tear & n.a. & partially \\
\hline 
9 & used & convex & with tear & n.a. & partially \\
\hline 
10 & used & convex & no tear & n.a. & fully \\
\hline 
11 & used & convex & with tear & n.a. & fully \\
\hline 
\end{tabular}
\end{table}

\begin{table}[htb]
\centering
\caption{Contradicting result paths between level 1 and 2}\label{tab:Konfliktkategorien}
\begin{tabular}{|c|c|c|c|}
\hline 
\textbf{Hierarchy level} & \textbf{1} & \multicolumn{2}{c|}{\textbf{2}} \\
\hline 
\textbf{No.} & \textbf{Use condition} & \textbf{Flap profile} & \textbf{Flap tear} \\
\hline
\hline
1 & new & any & with tear \\
\hline 
2 & new & concave & any \\
\hline 
3 & new & convex & any \\
\hline 
\end{tabular} 
\end{table}

A transfer learning approach is pursued to ensure a high degree of generalizability for the individual networks of the partial decisions with limited domain-specific data sets. Networks that have already been pre-trained on extensive image databases are adapted to the specific classification task by adjusting the output layer. In this work, EfficientNetV2 serves as backbone. This architecture offers high feature extraction efficiency with a comparatively low computational load. Another advantage is its availability in different scales (S, M, L) with identical feature vector dimensions, which allows for flexible exchange of model sizes without structural adjustments to the downstream architecture \citep{Tan.2021}.

\section{Data Generation and Implementation}\label{sec:Implementation}
\subsection{Image acquisition}
\begin{figure}[htb]
\centering
\includegraphics{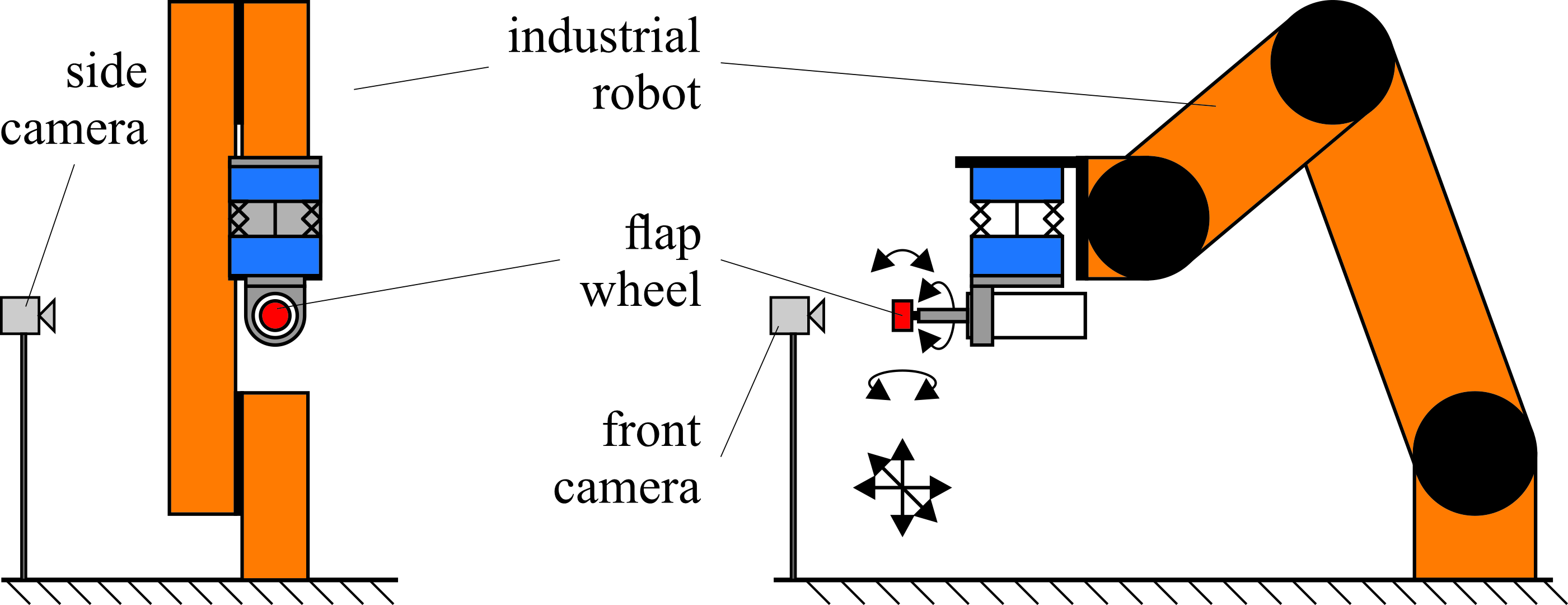}
\caption{Schematic setup for image acquisition. Left: Radial perspective. Right: Axial perspective.\label{fig:SetupDatengenerierung}}
\end{figure}

For image acquisition a robotic setup was used. The setup consists of an industrial robot carrying the spindle with flap wheel. Next to the robot, two cameras were mounted, in front of which the robot positions the flap wheel. The position and orientation of the flap wheel was adjusted randomly by $\pm5$ mm and $\pm5$ degrees for each image in order to vary the image composition and cover minor tolerances when calibrating the tool. In addition, the flap wheel was rotated randomly by the spindle, but each image was taken with the tool at a standstill. No dedicated lighting was used, only ambient light in a test hall was utilized. However, controlled lighting conditions will be ensured in the future. The schematic setup for image acquisition is shown in Figure \ref{fig:SetupDatengenerierung}.

In total, 13,240 images of 105 different flap wheels were captured, covering different flap wheel dimensions as well as wear patterns described in section \ref{sec:WearAnalysis}. 50\% of these images were taken from a radial perspective and 50\% from axial perspective. Before image acquisition, the tool wear condition was determined manually, each image of the specific tool was then labeled automatically. For each model of the hierarchical classification, suitable images were selected and divided into training, validation and test sets, so that each tool is present in only one of these subsets. However, according to the advantages of hierarchical classification approaches, images of a specific tool can be used for multiple models, e.g. usage classification and tear classification, avoiding data fragmentation. Table \ref{tab:DatasetDetails} gives details about the dataset composition, including the number images as well as the number of flap wheels for each class during training and testing.

\begin{table}[htb]
    \centering
    \caption{Dataset composition}
    \label{tab:DatasetDetails}
    \begin{tabular}{|c|c|cc|cc|}
    \hline
        \multirow{2}{*}{\textbf{Level}} & \multirow{2}{*}{\thead{\textbf{Model}\\ \textbf{(classes)}}} & \multicolumn{2}{c|}{\textbf{Training/Validation}} & \multicolumn{2}{c|}{\textbf{Test}} \\
        
         &  &  No. tools &  Total images&  No. tools &  Total images \\
         \hline
         1&  \thead{\textbf{Usage condition}\\ \textbf{(new/used)}}& 5/43 & 3700 & 7/14 & 1500\\
         2&  \thead{\textbf{Flap tear}\\ \textbf{(with/without)}}& 11/31 & 3020 &  12/17 & 1153\\
         2&  \thead{\textbf{Flap profile}\\ \textbf{(rect/concave/convex)}}& 26/33/22 & 2720 &13/16/24  & 3564\\
         3&  \thead{\textbf{Concavity}\\ \textbf{(fully/partially)}} & 14/19 &  720& 4/4 & 440\\
         3&  \thead{\textbf{Convexity}\\ \textbf{(fully/partially)}}& 12/10 & 1400 & 4/4 & 380\\
         \hline
    \end{tabular}
\end{table} 

\subsection{Data preprocessing and model training}
\begin{figure}[htb]
\centering
\includegraphics{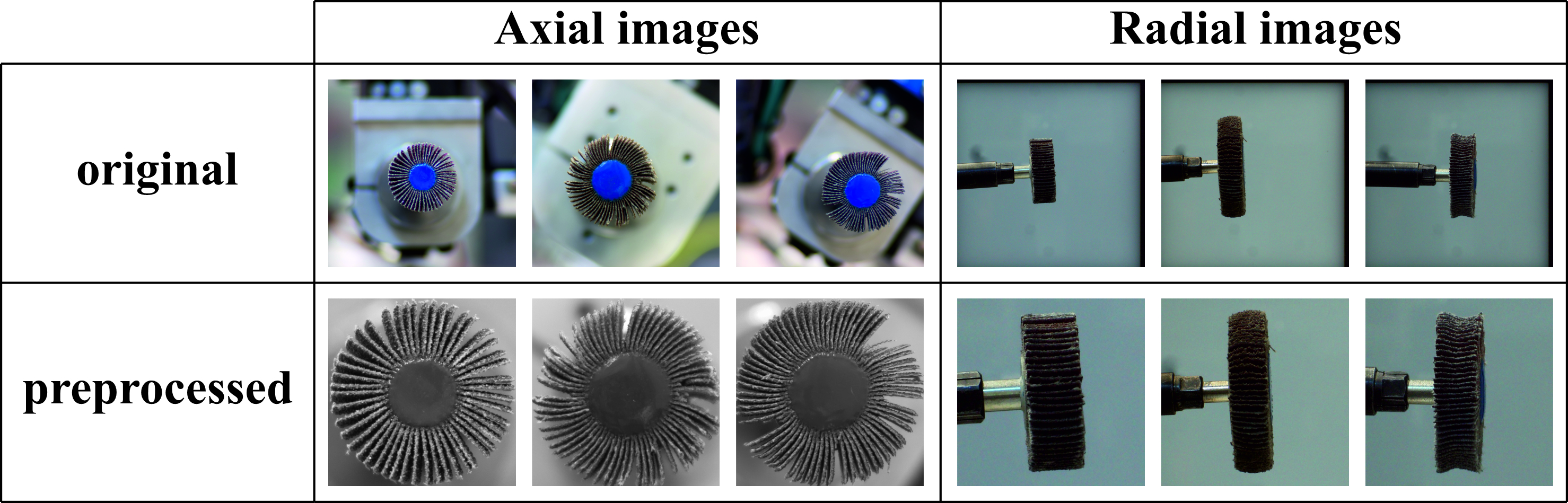}
\caption{Examples of captured images and after preprocessing\label{fig:DatasetExamples}}
\end{figure}

The original images were preprocessed by reducing noise and cropped, so that the resulting image corresponds to the flap wheel dimensions and excludes most of the background. The image cropping was determined automatically based on the detected flap wheel edges. Figure \ref{fig:DatasetExamples} shows exemplary dataset samples, original images as well as corresponding preprocessed images. Then, images were resized to 512x512 pixel and normalized channelwise. In addition, axial images were converted to grayscale, as colors are less relevant for flap tear detection. Additional augmentations such as image rotation, random brightness and contrast adjustment were applied during training.

The hierarchical classification and each model were implemented using PyTorch. For the tear classification model EfficientNetV2-L was used to increase feature extraction and selectivity, while the other models used EfficientNetV2-S. Each model was trained separately and the model validation performance was tracked. AdamW optimizer was applied. The parameters were selected based on research and best practices and amounted to $1e-4$ for both learning rate and weight decay. After training was completed, the model of an epoch with the best validation accuracy was selected.

\section{Results and discussion}\label{sec:Results}
In this section, the results of the classification approach are presented and discussed. First, each model is evaluated separately, then the overall performance of the hierarchical classification is analyzed. For each model, test samples were classified, and the class with the highest probability determines the final prediction. To quantify this classification performance, metrics such as accuracy, precision, recall and F1-score are used. In addition, ROC-curves and Grad-CAMs were further examined.

\subsection{Usage condition classification}
To evaluate the usage classification network, 1,500 images were used. Table \ref{tab:KonfusionsmatrixNutzungszustand} shows the corresponding confusion matrix and performance metrics.

\begin{table}[htb]
\centering
\caption{Confusion matrix and performance metrics of usage condition classification}\label{tab:KonfusionsmatrixNutzungszustand}
\begin{tabular}{|l|c|c||c|c|c|}
\hline
\diagbox{\textbf{True Class}}{\textbf{Prediction}} & \textbf{new} & \textbf{used} & \textbf{Precision} & \textbf{Recall} & \textbf{F1-score}\\ 
\hline
\textbf{new} & 458 & 2    &0.960 &0.996 &0.978\\ 
\hline
\textbf{used} & 19 & 1021 &0.998 &0.982 &0.990\\ 
\hline
\end{tabular}
\end{table}

\begin{figure}[htb]
\centering
\includegraphics{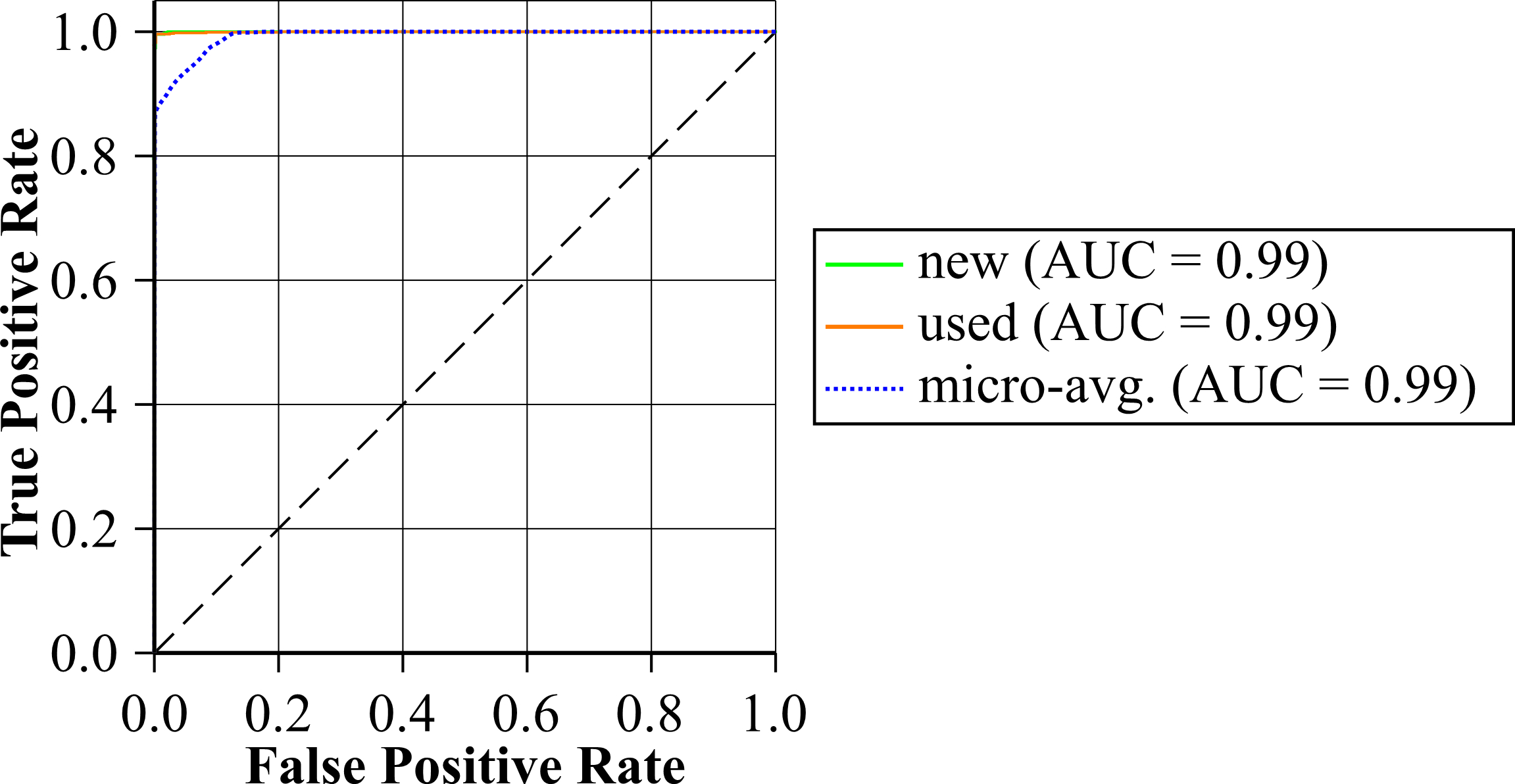}
\caption{Receiver Operating Characteristic (ROC) for the usage condition classification model\label{fig:ROC_Wear}}
\end{figure}

Overall, a good accuracy of $J_{usage} = 0.986$ and macro F1-score of $F1_{usage} = 0.983$ were achieved, indicating a very good classification performance. The ROC-curves also indicate that class imbalance does not seem to influence the model. The robustness of the Level 1 classification is further illustrated by the ROC curves in Figure \ref{fig:ROC_Wear}. The plot indicates near-perfect separation between 'new' and 'used' states, yielding an Area Under the Curve (AUC) of approximately 0.999 for both classes.

A deeper analysis revealed that incorrect predictions are generally made with lower confidence ($\overline{\hat{y}}_{false}=0.89$) compared to the overall average confidence ($\overline{\hat{y}}=0.97$). Based on the confusion matrix, the greatest shortcoming is in the miscategorization of worn tools, which are falsely predicted as new. Analyzing the false classifications, the majority of misclassifications originated from a rectangular worn flap wheel, which has similar characteristic fringes on the radial edge of the flap as new ones. As these fringes usually detach during use, classifying flap wheels after a certain use time might circumvent this issue. In practice, the tool condition can also be determined by tracking whether they have been used.

\subsection{Flap profile classification}
The flap profile is determined in level 2 and 3 of the hierarchical classification, therefore both are evaluated in this section. For testing the general flap profile classification (level 2), 3564 images were used, of which 163 were misclassified. This resulted in an accuracy $J_{profile}$ of 0.954 and $F1_{profile}$ of 0.954. All rectangular flap wheels have been classified correctly. Concave ones received a misclassifications rate of about 11\% instead, which will not be sufficient for industry deployment.

\begin{figure}[htb]
\centering
\includegraphics{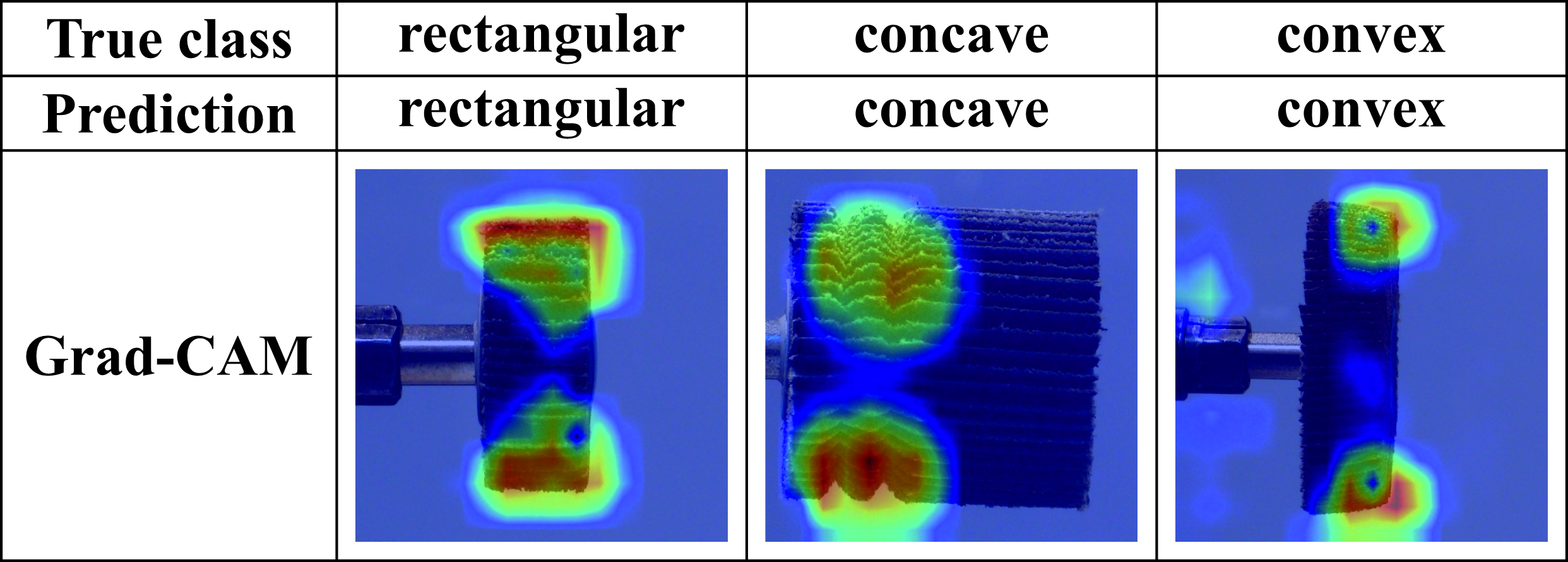}
\caption{Grad CAM overlays of correct flap profile classifications\label{fig:GradCAM_Flap_Profile}}
\end{figure}

Grad-CAM overlays of true positive samples are shown in Fig. \ref{fig:GradCAM_Flap_Profile} and reveal that for tools assigned to the “convex” class strong activations can be seen in the tapered areas of the flap. Convex worn tools also exhibit these zones, with the difference that they are arranged in the opposite direction. The specimen identified as rectangular shows activation at the transition between the radial outer edge and the front side of the flap wheel.

\begin{figure}[htb]
\centering
\includegraphics{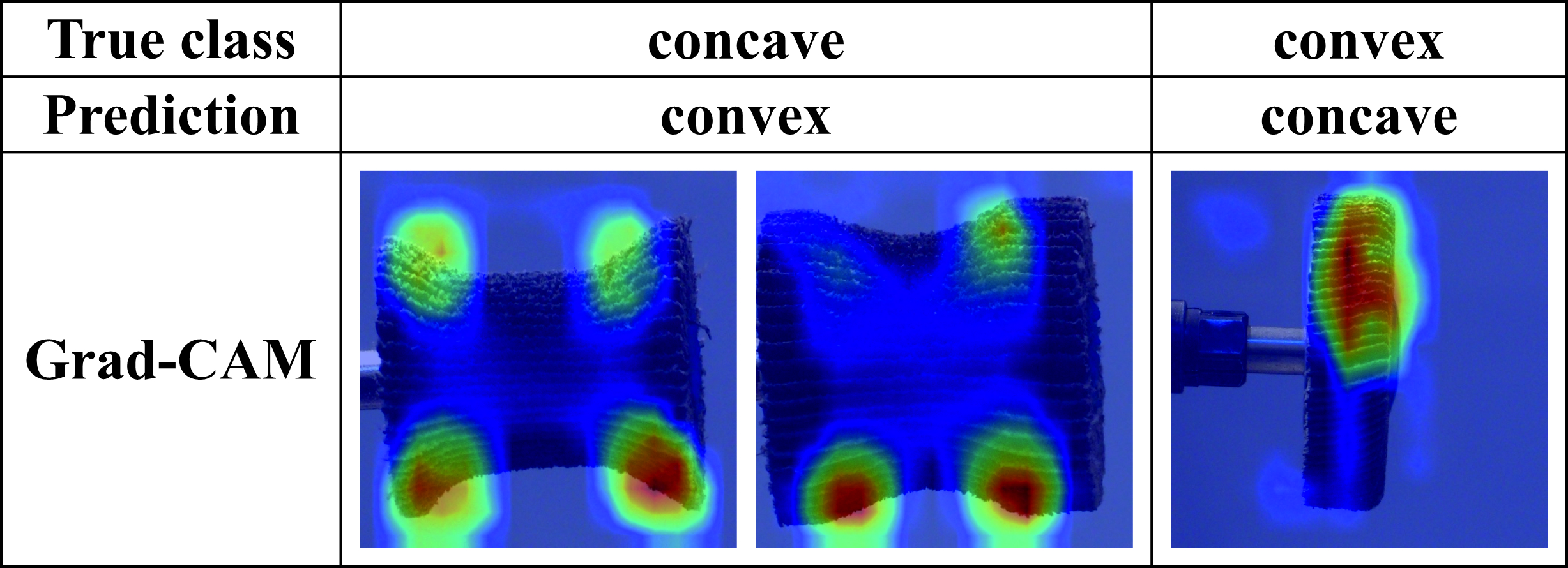}
\caption{Grad CAM overlays of false flap profile classification\label{fig:GradCAM_Flap_Profile_Falses}}
\end{figure}

However, incorrect classifications can be caused by the rotational symmetry of flap wheels. As seen on the left in Figure \ref{fig:GradCAM_Flap_Profile_Falses}, the model correctly uses the upper and lower tapered flap edge as features to determine the shape of the flap grinding wheel. However, stronger activations can be seen in the lower half of the flap wheel, suggesting that the flap contour may have been recognized as convex instead of concave. Also, irregular flap inclination due to tool manufacturing tolerances seem to influence the classification, as seen on the right in Figure \ref{fig:GradCAM_Flap_Profile_Falses}. These observations suggest that vertical flip augmentation could have helped to ensure that the upper and lower flap edges were given equal consideration in the decision, even if this meant losing the direction of rotation of the flaps.

\begin{table}[htb]
\centering
\caption{Confusion matrix and performance metrics of flap profile classification (level 2)}\label{tab:KonfusionsmatrixLamellenprofil}
\begin{tabular}{|l|c|c|c||c|c|c|}
\hline
\diagbox{\textbf{True class}}{\textbf{Prediction}} & \thead{\textbf{rect-}\\ \textbf{angular}} & \textbf{concave}& \textbf{convex} & \textbf{Precision} & \textbf{Recall} & \thead{\textbf{F1-}\\ \textbf{score}} \\ 
\hline
\textbf{rectangular}& 1165&0   & 0   & 0.955&1.000 &0.977 \\ 
\hline
\textbf{concave}    & 40  &1212& 107 &0.999&0.892&0.943\\ 
\hline
\textbf{convex}     & 15  & 1  & 1024&0.905&0.985&0.943\\ 
\hline
\end{tabular}
\end{table}

Moving on to level 3, both models for classifying the severity of concave and convex flap shapes achieved excellent accuracies of $J_{concave} = 0.993$ and $J_{convex} = 0.950$, as well as macro F1-scores of $F1_{concave} = 0.993$ and $F1_{convex} = 0.948$, indicating the concave severity model outperformed the convex severity model slightly. Analyzing the confusion matrices, particularly for convex profiles (Table \ref{tab:KonfusionsmatrixLamellenprofil_konvex}), reveals a directed error pattern. While the model achieves perfect recall for 'partially' worn tools, it exhibits a tendency to misclassify 'fully' worn tools as 'partially' worn (19 instances). This suggests that the transition from partial to full convexity is fluid.

\begin{table}[H]
\centering
\caption{Confusion matrix and performance metrics of concave profile severity classification (level 3)}\label{tab:KonfusionsmatrixLamellenprofil_konkav}
\begin{tabular}{|l|c|c||c|c|c|}
\hline
\diagbox{\textbf{True Class}}{\textbf{Prediction}} & \thead{\textbf{fully}\\ \textbf{concave}} &  \thead{\textbf{partially}\\ \textbf{concave}} & \textbf{Precision} & \textbf{Recall} & \thead{\textbf{F1-}\\ \textbf{score}}\\ 
\hline
\textbf{fully concave}      & 157   & 3    &1.000&0.981&0.991 \\ 
\hline
\textbf{partially concave}  & 0     & 280  &0.989&1.000&0.995\\ 
\hline
\end{tabular}
\end{table}

\begin{table}[H]
\centering
\caption{Confusion matrix and performance metrics of convex profile severity classification (level 3)}\label{tab:KonfusionsmatrixLamellenprofil_konvex}
\begin{tabular}{|l|c|c||c|c|c|}
\hline
\diagbox{\textbf{True Class}}{\textbf{Prediction}} & \thead{\textbf{fully}\\ \textbf{convex}} & \thead{\textbf{partially}\\ \textbf{convex}} & \textbf{Precision} & \textbf{Recall} & \thead{\textbf{F1-}\\ \textbf{score}}\\ 
\hline
\textbf{fully convex} & 141 & 19 &1.000&0.881&0.937 \\ 
\hline
\textbf{partially convex} & 0& 220 &0.921&1.000&0.959\\ 
\hline
\end{tabular}
\end{table}

\begin{figure}[htb]
\centering
\includegraphics{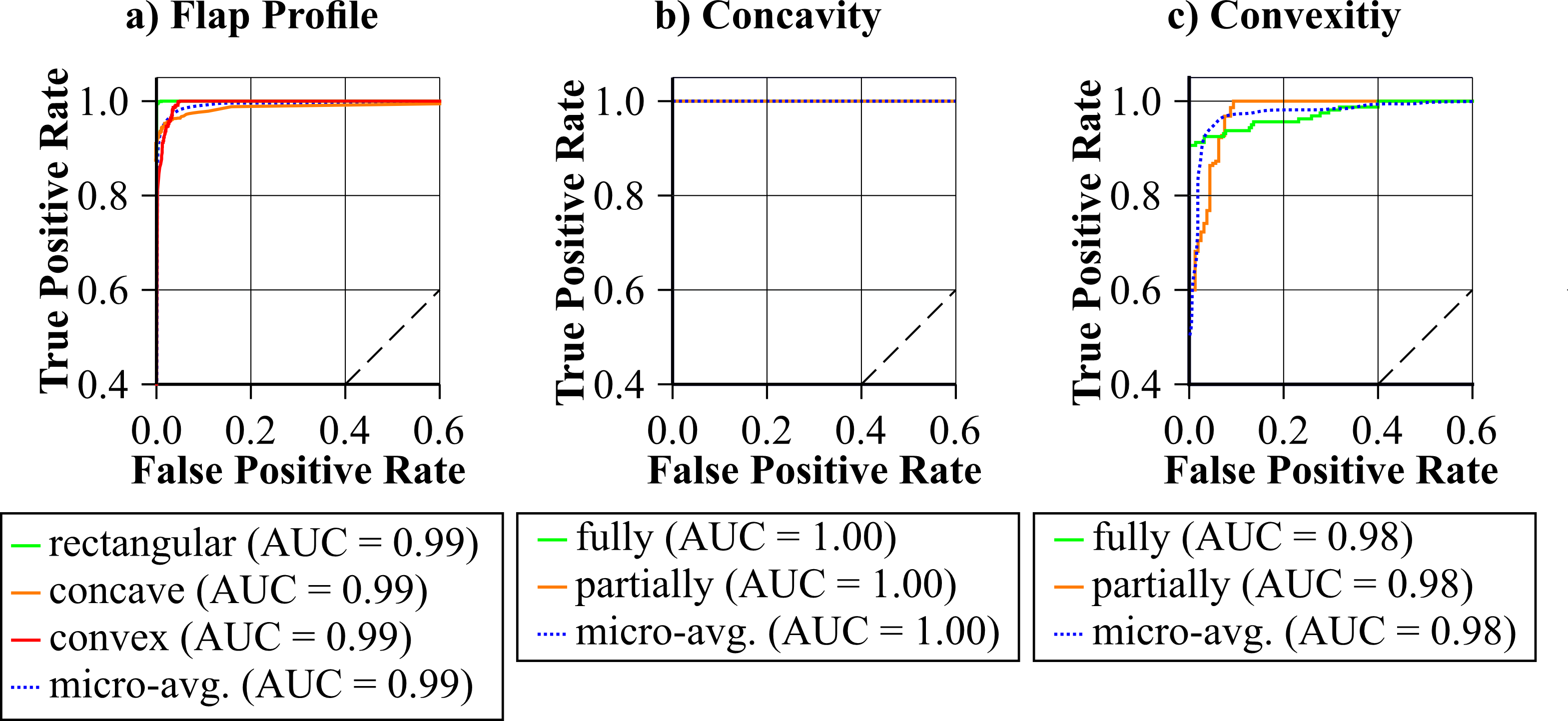}
\caption{Receiver Operating Characteristic (ROC) for the shape classification models. a) Flap profile classification (level 2). b) Concave severity classification (level 3). c) Convex severity classification (level 3)\label{fig:ROC_Shape}}
\end{figure}

Figure \ref{fig:ROC_Shape} illustrates the Receiver Operating Characteristic (ROC) curves for the geometric profile classification across hierarchy levels 2 and 3. The curves demonstrate the exceptional separability of the classes, with the Area Under the Curve (AUC) approaching or reaching the ideal value of 1.00. In Figure \ref{fig:ROC_Shape} a), the general profile classification (Level 2) exhibits a steep initial ascent for all classes (rectangular, concave, convex), resulting in AUC values of 0.99. This indicates that the model can maintain a high true positive rate while keeping false positives negligible, effectively validating the distinctiveness of the learned geometric features.

The severity assessment in Level 3 shows a slight divergence in performance depending on the wear type. While the differentiation between fully and partially concave profiles is perfectly solved with an AUC of 1.00 (Figure \ref{fig:ROC_Shape} b)) , the convex classification (Figure \ref{fig:ROC_Shape} c)) shows a minor performance drop. This correlates with the findings from the confusion matrix (Table \ref{tab:KonfusionsmatrixLamellenprofil_konvex}), indicating that the visual transition from partial to full convexity is harder to discretize than for concave wear patterns. Nevertheless, the high AUC values ($ \ge 0.98$) across all classes demonstrate that the models remain unbiased despite a certain class imbalances in the dataset, expecting that less frequent wear severities are detected with the same reliability as dominant patterns.

\subsection{Flap tear classification}
Based on the confusion matrix in Table \ref{tab:KonfusionsmatrixLamellenausrisse}, 72 of 1,153 test samples were misclassified, giving an accuracy $J_{tear}$ of 0.938 and $F1_{tear}$ of 0.935. The mean confidence was $\overline{\hat{y}} = 0.92$, while misclassifications were done with a confidence of $\overline{\hat{y}}_{false} = 0.76$. The differences indicate a meaningful separation of confidence levels and offers potential for targeted verification of predictions with lower probability, in particular to increase sensitivity without significantly compromising high precision.

\begin{figure}[htb]
\centering
\includegraphics{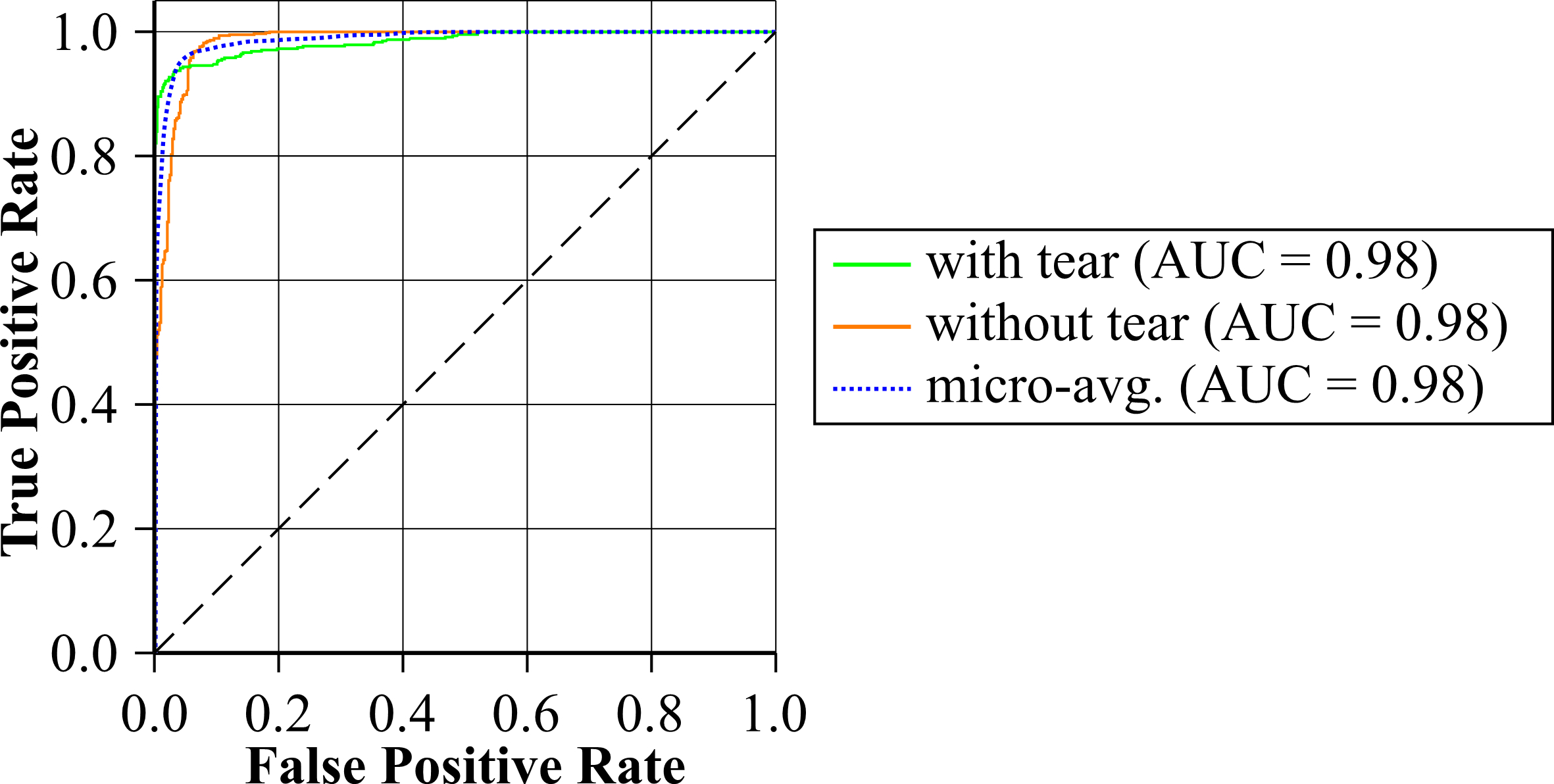}
\caption{Receiver Operating Characteristic (ROC) for the tear classification model\label{fig:ROC_Tear}}
\end{figure}

The ROC analysis (see Figure \ref{fig:ROC_Tear}) yields an AUC of 0.98, confirming the model's high sensitivity to flap tears. This performance indicates that the system can reliably detect flap tears with only a low risk of false positives.

\begin{table}[htb]
\centering
\caption{Confusion matrix and performance metrics of flap tear classification}\label{tab:KonfusionsmatrixLamellenausrisse}
\begin{tabular}{|l|c|c||c|c|c|}
\hline
\diagbox{\textbf{True Class}}{\textbf{Prediction}} &  \thead{\textbf{with}\\ \textbf{tear}} &  \thead{\textbf{without}\\ \textbf{tear}}& \textbf{Precision} & \textbf{Recall} & \thead{\textbf{F1-}\\ \textbf{score}}\\ 
\hline
\textbf{with tear} & 419 & 61 &0.974&0.873& 0.921\\ 
\hline
\textbf{without tear} & 11 & 662&0.916&0.984& 0.948\\ 
\hline
\end{tabular}
\end{table}

\begin{figure}[htb]
\centering
\includegraphics{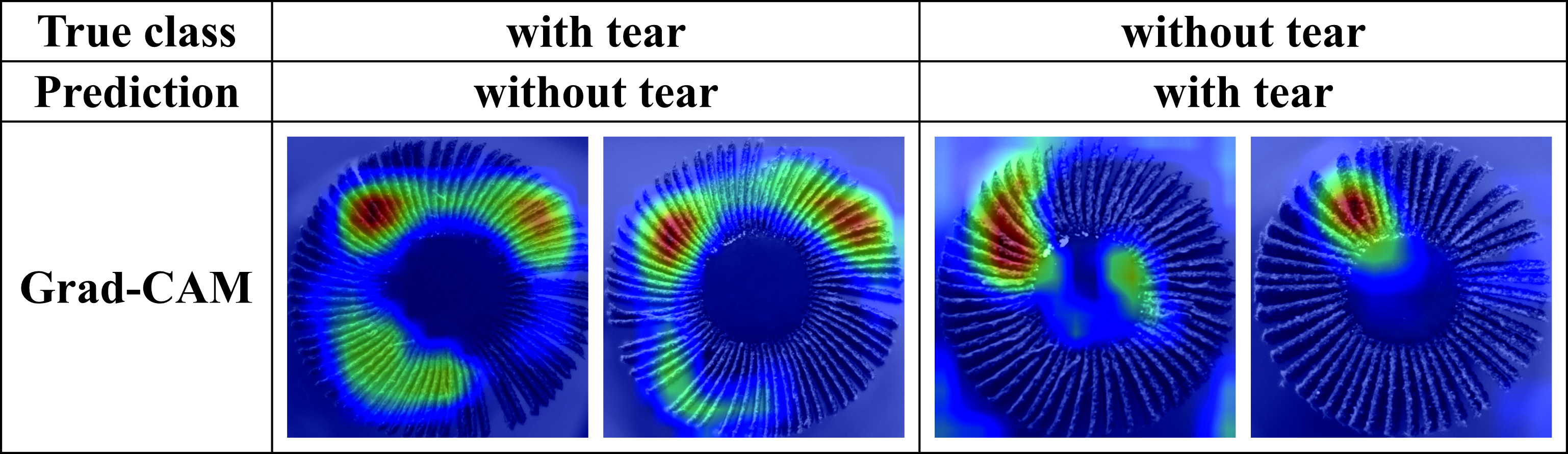}
\caption{Grad CAM overlays of false flap tear classifications\label{fig:GradCAM_Flap_Tear}}
\end{figure}

The Grad-CAM overlays in Figure \ref{fig:GradCAM_Flap_Tear} show that some activations leading to incorrect classification (false no tear) do not originate from the areas with flap tear. In contrast, it can be observed that samples incorrectly assigned to the class with tear (false tear) show flap wheels with irregular spacing between the flaps. This suggests performance may be improved by providing more training samples for both classes, but irregular flap spacing. Additional flap wheel images with single flap tears may also improve selectivity, as the model tends to miss those more often than multiple tears next to each other.

Another starting point for improvement would be to exploit the flap properties to optimize image capture and assist the network. Since the flaps extend radially during rotation, unlike when stationary, the flap straightness and spacing is expected to become more uniform. However, since higher rotation speeds are required to extend the flaps, good-quality images require more sophisticated image acquisition, e.g. need for a global shutter camera, short exposure time and likely additional lighting.

\subsection{Hierarchical classification}
For testing of the hierarchical classification, nine flap wheels were used, which have not been included in previous training or testing of individual models. 40 radial and 40 axial images of each flap wheel were captured. In a classification run, one radial and one axial image of each tool were selected randomly and passed to the hierarchical classification. This resulted in 40 runs per flap wheel and 360 total runs.

45 misclassifications occurred, giving an accuracy $J_{hierarchy}$ of 0.875, which is located within the expected overall accuracy interval between 0.838 and 0.882, depending on which models are actually used (see Equations \ref{eq:Jrect}, \ref{eq:Jconcave} and \ref{eq:Jconvex}). Comparatively low to the individual models, this shows the inherent weakness of a hierarchical classification: The overall accuracy is the product of each submodel accuracy.

\begin{equation}\label{eq:Jrect}
    J_{hierarchy, rect} = J_{usage} \cdot J_{tear} \cdot J_{profile} = 0.882
\end{equation}
\begin{equation}\label{eq:Jconcave}
    J_{hierarchy, concave} = J_{usage} \cdot J_{tear} \cdot J_{profile} \cdot J_{concave}= 0.876
\end{equation}
\begin{equation}\label{eq:Jconvex}
    J_{hierarchy, convex} = J_{usage} \cdot J_{tear} \cdot J_{profile} \cdot J_{convex}= 0.838
\end{equation}

Analyzing the misclassifications, 22 resulted from the usage condition model (level 1) and 23 from the flap tear model (level 2). Measures for increasing the hierarchy performance can be derived from these results. In case of contradictions in the hierarchy, additional examinations should be triggered. In this case, this may have covered 4 misclassifications (9\% of errors). If the confidence of these two submodels would be considered and a suitable confidence threshold applied, uncertain decisions and additional errors could be caught. For the usage condition model, a confidence threshold of $\hat{y} = 0.91$ would have detected 11 of the 22 misclassifications, while 10 correct classifications would have been checked again. For the tear classification model, a threshold of $\hat{y} = 0.79$ would have detected 11 more misclassifications, with 9 correct classifications rechecked.

By implementing the conflict and confidence mechanisms, and assuming the re-examination yields a correct result, a theoretical accuracy between 93.6\% and 94.7\% could be expected, depending on whether the conflicts are already included in the threshold errors or are found additionally. The performance is expected to improve further when several individual classifications of multiple tool images are combined into an ensemble classification.

\section{Conclusion and outlook}\label{sec:Conclusion}
This paper introduced an image-based hierarchical classification approach to determine the geometric wear condition of abrasive flap wheels. By utilizing multiple networks, the wear state was decomposed into different features and decisions, and the wear was refined in three levels from coarse to fine. This decomposition not only reduced the complexity for individual classifiers but also enabled logical consistency checks, preventing contradictory predictions such as declaring a new tool with flap tear.

The implemented networks were based on EfficientNetV2 and individually achieved high accuracies and F1-scores above 0.9 respectively. In some cases, however, performance is yet too low for autonomous use. Grad-CAM analysis confirmed that the models focus on physically relevant wear features such as flap profile contours or flap gaps. To further minimize error propagation in the hierarchical chain, future work could implement dynamic confidence thresholds to trigger re-evaluation of uncertain predictions, as well as using multiple images of the same tool to ensemble a final prediction.

Future work will link the proposed classification to explicit wear measurement to generate a holistic wear picture. In this pipeline, the wear classification stage could serve as a robust validation mechanism for downstream quantification methods. While generic image processing filters can extract contours, combining them with an upstream classification minimizes ambiguity. For instance, the classification can provide a semantic context (e.g., 'concave profile') that verifies the analytically derived geometry. If the measurement algorithm detects a curvature that aligns with the classified wear type, the confidence in the result is significantly increased. Additionally, this a priori knowledge allows for the targeted selection of fitting algorithms (e.g., circular arc fitting versus linear regression), ensuring that the most relevant geometric parameters for adaptive process compensation are prioritized. This cross-validated data forms a solid basis for deciding how to proceed with the flap wheel—whether to compensate wear, dress or finally replace the tool. Consequently, integrating classification and measurement into an automation system contributes to increasing process autonomy and reliability.


\section*{Acknowledgements}

\begin{wrapfigure}[3]{r}{0.2\linewidth}
\raisebox{-40pt}[0pt][0pt]{
\includegraphics[width=0.9\linewidth]{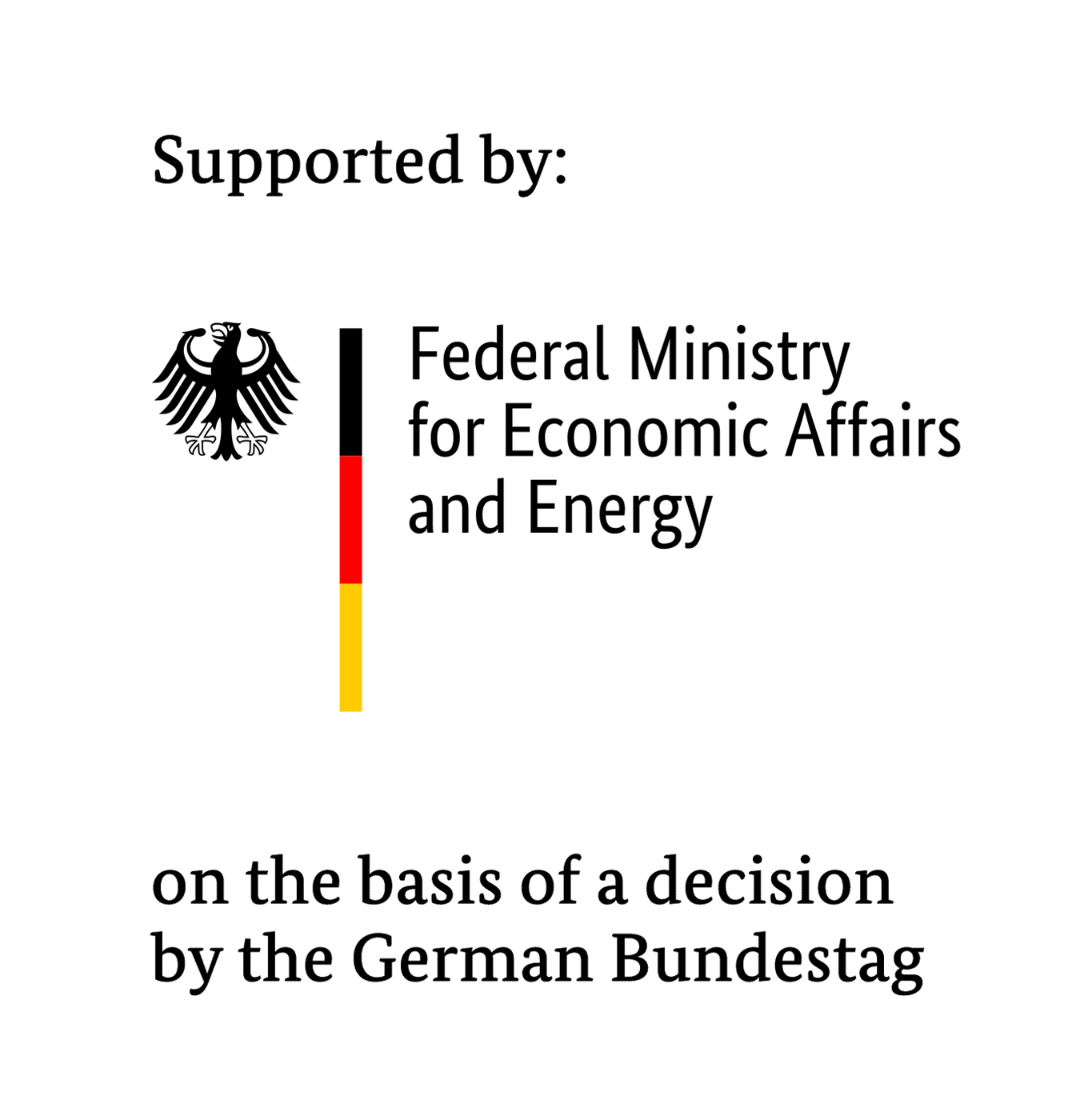}}
\end{wrapfigure}

\par
\par

The research leading to these results received funding from the Federal Ministry for Economic Affairs and Energy (BMWE) in the project LEGATO under the Grant Number 20Y2205C as part of the Federal Aeronautical Research Programme LuFo VI-3.

\section*{Authors' contributions} \textbf{Falko Kähler}: Conceptualization, Methodology, Writing - original draft. \textbf{Maxim Wille}: Conceptualization, Methodology, Writing - review. \textbf{Ole Schmedemann}: Review. \textbf{Thorsten Schüppstuhl}: Funding acquisition.

\bibliographystyle{unsrtnat} 
\bibliography{literature}    

\end{document}